# Gesture Matters: Pedestrian Gesture Recognition for AVs Through Skeleton Pose Evaluation

Alif Rizqullah Mahdi
*Institute for Transport Studies*
*University of Leeds*
Leeds, United Kingdom
tsarm@leeds.ac.uk

Mahdi Rezaei
*Institute for Transport Studies*
*University of Leeds*
Leeds, United Kingdom
m.rezaei@leeds.ac.uk

Natasha Merat
*Institute for Transport Studies*
*University of Leeds*
Leeds, United Kingdom
n.merat@its.leeds.ac.uk

*Abstract*—Gestures are a key component of non-verbal communication in traffic, often helping pedestrian-to-driver interactions when formal traffic rules may be insufficient. This problem becomes more apparent when autonomous vehicles (AVs) struggle to interpret such gestures. In this study, we present a gesture classification framework using 2D pose estimation applied to real-world video sequences from the WIVW dataset. We categorise gestures into four primary classes (Stop, Go, Thank & Greet, and No Gesture) and extract 76 static and dynamic features from normalised keypoints. Our analysis demonstrates that hand position and movement velocity are especially discriminative in distinguishing between gesture classes, achieving a classification accuracy score of 87%. These findings not only improve the perceptual capabilities of AV systems but also contribute to the broader understanding of pedestrian behaviour in traffic contexts.

*Keywords—pedestrian gestures, autonomous vehicle, human-machine interaction, gesture recognition, skeleton pose analysis*

## I. Introduction

As cities evolve toward intelligent transportation systems, implementing and deploying autonomous vehicles (AVs) is becoming increasingly inevitable. While traffic behaviour and regulations are generally regulated by legal frameworks, particularly those that define right-of-way, situations may arise where road users need to waive their priority [1]. In such cases, a critical yet often overlooked factor is the role of the day-to-day, non-verbal communication between pedestrians and human drivers. This form of communication exists as a set of informal rules that direct and influence real-world traffic situations [2]. Among these, gestures are an important form of explicit communication that enables road users to negotiate their intentions and ensure mutual understanding [3].

For example, a study in Qatar shows drivers have a 2.8 times higher yielding rate when pedestrians gesture towards them [4]. Rasouli et al. also stated that, almost 90% of the time, pedestrians show some form of micro gestures, including head and eye orientation before crossing, and among them, only a very small percentage (<2%) use explicit gestures [5] Other studies have also suggested that explicit gesturing is relatively rare in daily traffic scenarios [6, 7]; however, it becomes necessary in deadlock situations where formal rules have no resolution and the solution can only be established through non-verbal cues, such as head orientation or hand gestures [5].

In human-driven traffic, shared awareness between road users is critical in developing trust and ensuring safety. It involves not only perception but also the integration of domain-specific knowledge and causal reasoning of pedestrian behaviours and intentions [8]. Human drivers can often interpret non-verbal cues to anticipate a pedestrian's action or intention. However, AVs lack the ability to perceive and understand these social signals with the same capacity as humans do [9]. They rely primarily on predefined rules and sensor inputs that may not capture the variation and uncertainty of human behaviour. This gap could become troublesome in situations requiring two-way communication and negotiation, such as during deadlocks or unsignalised crossings. Without the ability to interpret and respond to such cues, AVs have the risk of becoming overly cautious, unsafe, or even unresponsive in complex urban environments [3].

Recent research findings further emphasised the importance of gestures in pedestrian-AV communication. A Virtual Reality (VR) study also found that pedestrians use gestures in 70% of interactions to communicate their intent when encountering AVs [10]. Hafeez et al., through a multilingual survey, support this finding by mentioning that pedestrians often rely on non-verbal cues to express their intentions [11].

Despite the number of studies carried out on the subject of pedestrian gestures, there is a lack of a specifically tailored dataset for pedestrian gesture recognition. Existing datasets tend to be limited in scope, constrained to specific gesture categories, sensing modalities, or simulated settings. For instance, the GLADAS dataset focuses on pedestrian gesture understanding using front-view simulated data [12], while the TCG dataset provides traffic police gesture recognition using 3D body pose data [13]. Although these datasets contribute to broad insights for pedestrian gesture recognition, without dedicated real-world data of pedestrian gestures, it would be difficult to compare and generalise the results of the research findings.

Although pedestrian gesture datasets remain scarce, a recent publication by Brand and Schmitz made a significant advancement for pedestrian gesture recognition by collecting real-world pedestrian gestures, known as the WIVW dataset [1]. Their work compares gesture expressiveness across virtual and physical environments, leaving an identifiable gap for computational analysis of these gestures. This paper aims to bridge this gap by proposing a computer vision pipeline to extract and interpret skeletal poses and motion dynamics from the WIVW videos. We introduce a four-class pedestrian gesture taxonomy, derived through a review of relevant literature and existing gesture-related datasets. Our work provides an advancement in standardising the gesture categorisation and provides a framework for evaluating pedestrian gestures in real-world traffic contexts.

## II. Gesture Taxonomy

### A. Gesture Classification

Pedestrian gestures can have different types and meanings across regions and cultures [14]. However, there are some similarities between these variations of gestures, which are commonly used for pedestrian-to-vehicle communication. For example, a stop signal using an L-bent hand gesture in Beijing

This work is funded by the UK Research and Innovation (UKRI) through Reference Numbers EP/W524372/1 and 2887445.

[15] has a similar pose to a stop gesture mentioned in Brand and Schmitz's work [1]. Moreover, some Chinese traffic police gestures have similar movements and poses to commonly used pedestrian gestures, such as for "Stop" and "Move Straight" [16]. We have defined these commonly used gestures based on our findings throughout multiple literatures, shown in Table 1.

TABLE 1 TYPES OF GESTURE AND MEANING

| Gesture | Position | Meaning | Reference |
|---|---|---|---|
| Raising a hand | Overhead | Asking to stop or pull over | Defined by [1, 15], also used by traffic police [16] |
| Extending an arm and/or hand | Chest-level | Asking to stop | Defined by [1, 4, 15], used by traffic police [16], and in a simulated study [12] |
| Extending and waving arm(s) vertically | Chest-level and hip-level | Asking to stop or slow down | Defined by [1], similar to straight-level pose in [15] |
| Raising a hand | Head-level | Thanking or greeting | Defined by [1, 15] |
| Waving hand | Near the head and shoulders | Greeting | Defined by [1] |
| Thumbs up | Chest-level | Thanking or greeting | Defined by [1, 15] |
| Swinging arm side-to-side | Chest-level and hip-level | Giving way or waiving | Defined by [1], used by traffic police [16], and in a simulated study [12] |

Based on the findings above, we derived a four-class taxonomy that defines the meaning of each gesture, i.e., "Stop", "Go", "Thank & Greet", and "No Gesture". These aggregated classes will help to improve the robustness of pedestrian gesture recognition across different regions and cultures. However, two main challenges persist:

1. Intra-class variation: A "stop" may involve a raised hand or extended arms
2. Inter-class ambiguity: Raising hand(s) could mean "stop" or "greeting", depending on the details of the movement

In order to mitigate these issues, we used post-processing methods to analyse the kinematic signatures (e.g., hand-to-hand distances, hand velocity) of gesture movements, as detailed in Section III.

This classification will be used as a reference for selecting video clips from the WIVW dataset [1]. Section I.B details the adaptation of the dataset to match our research scope, including the video extraction process.

*B. WIVW Dataset Overview*

The WIVW dataset [1] provides one of the most recent and comprehensive collections of pedestrian gestures, with 1000+ full high-definition (FHD) videos capturing gestures in experimental (indoor VR) and real-world settings. While originally designed to study gesture expressiveness across virtual and physical environments, we repurposed their videos for pedestrian gesture recognition by selecting clips that align with our four-class gesture classification, as discussed in the previous section.

To further ensure consistency with the topic of pedestrian-to-AV communication, we restricted our selection to outdoor front-view videos only, excluding indoor VR recordings and third-person perspectives. The indoor VR recordings have the participants wearing VR headsets, which may distort skeletal feature extraction, whereas the third-person perspective has suboptimal camera angles, where the camera is positioned lower than the pedestrian, and low resolution. From the 1000+ videos, we only procured videos from a pool of 361 cropped pedestrian-centric videos.

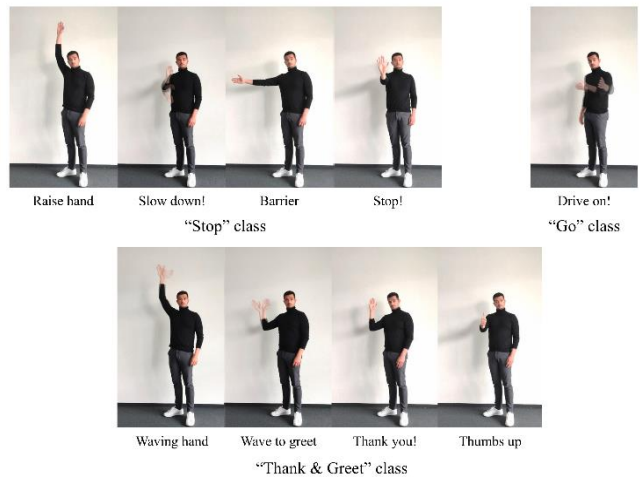

Fig. 1 Gesture illustration. (Courtesy of Thomas and Brand [1])

The dataset originally annotates nine gesture categories: i.e., "Stopping", "Waiving", "Warning", "Thanks", "Calling a Taxi", "Greeting", "Don't Understand", "Agree", and "Disagree". They are then further derived into 18 distinct motion patterns [1]. We excluded semantically ambiguous and context-specific gesture labels (e.g., "Don't Understand" and "Disagree") that do not align with our gesture classification. Furthermore, we only retained videos that match our movement definitions on Table 1. For instance, the "Stopping" ID in the WIVW dataset has a "Stop it!" motion (side-to-side arm wave) that does not correspond with our motion definition for the "Stop" class. This selection process yielded 180 videos: 53 for "Stop", 28 for "Go", 48 for "Thank & Greet", and 51 for "No Gesture", with the metadata summarised in Table 2 and the illustration of each gesture is shown in Fig. 1.

TABLE 2 WIVW DATASET MAPPING

| Class | WIVW Motion Pattern | Number of Videos |
|---|---|---|
| Stop | "Raise hand", "Slow down!", "Barrier", "Stop!" | 53 |
| Go | "Drive on!" | 28 |
| Thank & Greet | "Waving hand", "Wave to greet", "Thank you!", "Thumbs up" | 48 |
| No Gesture | "No Gesture" | 51 |

As mentioned previously, we utilised the WIVW's cropped pedestrian-centric videos, which were preprocessed by cropping the raw footage to bounding boxes around the detected pedestrians. This isolation guarantees that each frame contains only one pedestrian and eliminates any background clutter. We extracted the videos using OpenCV, maintaining the original resolution (1920x1080 pixels) and frame rate (60 FPS). However, the image sequences with the gesture-positive label include some non-essential frames that do not contribute to the gesture movement. Therefore, we further preprocessed the sequences to focus exclusively on the gesture-positive frames. The distribution of the duration of gestures for each class is shown in Fig. 2. By using only the gesture-positive frames, we can concentrate on verifying the characteristics of each gesture specifically.

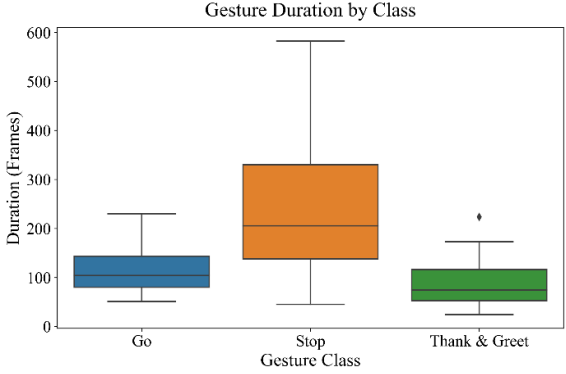

Fig. 2 Distribution of duration for each class.

III. METHODOLOGY

Our work aims to develop a pedestrian gesture recognition framework, specifically using skeleton point features, and provide an in-depth analysis of pedestrian gestures through post-processing techniques. A summary of the proposed framework is illustrated in Fig. 3. We utilise the MMPose [17] 2D pose estimation model, specifically the RTMPose-m [18], to extract the skeleton points from the image sequences. Compared to other pose estimation methods, such as MediaPipe [19], OpenPose [20], ViTPose [21], and ZoomNAS [22], MMPose provides a more robust and accurate skeleton pose estimation. One of the newest iterations of MMPose, the RTMW, has even achieved a real-time multi-person 2D and 3D whole-body pose estimation [23].

The 2D pose estimator provides 17 keypoints, consisting of $x$ and $y$ points that follow the COCO skeleton annotation [24] (0 for nose, 1 for left eye, 2 for right eye, etc.). An example estimated skeleton pose points overlayed over the original image is shown in Fig. 4. The resulting keypoints are still relative to the (0,0) point located at the upper leftmost point in the image. Therefore, normalisation is required to improve accuracy and aid in data post-processing.

We implemented a torso-based normalisation method to normalise the skeleton keypoints, inspired by Alzahrani et al.'s work [25]. Each keypoint will be compared with the torso size that is calculated based on the average Euclidean distances between the hips and shoulder, including cross distances between the left and right points. This formulation is represented by (1).

$$ts_t = \frac{\|\mathbf{p}_t^{LS}-\mathbf{p}_t^{LH}\|+\|\mathbf{p}_t^{LS}-\mathbf{p}_t^{RH}\|+\|\mathbf{p}_t^{RS}-\mathbf{p}_t^{LH}\|+\|\mathbf{p}_t^{RS}-\mathbf{p}_t^{RH}\|}{4} \quad (1)$$

where $ts_t$ is the torso size at frame $t$, $\mathbf{p}_t$ is the keypoint at frame $t$, and $LS$, $LH$, $RS$, $RH$ are left shoulder, left hip, right shoulder, and right hip points, respectively. Adding the cross distances will help in reducing the variance of bent torsos or other similar movements. Each keypoint $\mathbf{p}_t^i = (x_t^i, y_t^i)$ will be normalised relative to a centre point ($\mathbf{c}_t$) located in the body. Our work uses the mid-hip point as $\mathbf{c}_t$, as shown in (2).

$$\mathbf{c}_t = \frac{\mathbf{p}_t^{LH}+\mathbf{p}_t^{RH}}{2} \quad (2)$$

$$\hat{\mathbf{p}}_t^i = \frac{\mathbf{p}_t^i - \mathbf{c}_t}{ts_t}, \quad \forall i \in \{1,\ldots,17\} \quad (3)$$

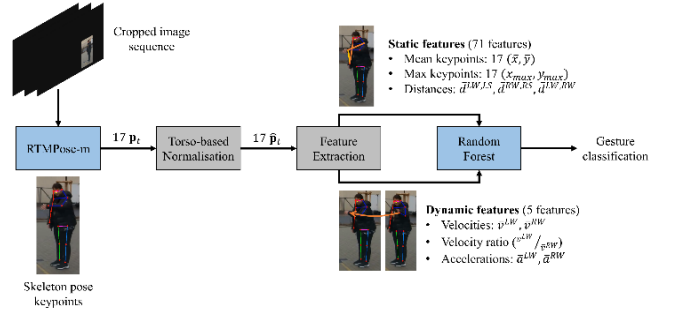

Fig. 3 Proposed framework.

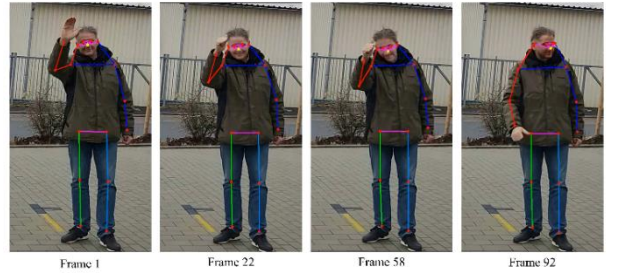

Fig. 4 Pose estimation example from the WIVW dataset.

The keypoints are then normalised based on the distance to $\mathbf{c}_t$, calculated using (3). This method ensures that the points are normalised using one parameter from the human body that rarely changes when pedestrians are gesturing, namely the torso. Other points of the body may change during movement. For example, if the total height is used as a reference, the value may change with limb or head motions.

Analysing gestures means analysing a sequence of actions or movements over a certain time [26]. While skeletal keypoints provide the pose estimation for each frame, gesture recognition requires analysing temporal evolution. We address this by extracting static and dynamic features from the keypoints. The following subsections discuss our feature engineering pipeline.

A. Static Feature

Static features capture pose configuration that can provide an estimation to distinguish gestures at individual timesteps. Building upon the normalised keypoints, we derived three static feature descriptors:

1. Mean keypoint position

The average position of keypoints could help provide an initial guess on the gesture, formulated by (4). For instance, the "Thank & Greet" class may have a higher average hand position compared to the "Go" class, due to the difference between hand-raising movements and arm-swinging movements.

$$\bar{x}^i = \frac{1}{t}\sum_{(t=0)}^{t} \hat{x}_t^i, \quad \bar{y}^i = \frac{1}{t}\sum_{(t=0)}^{t} \hat{y}_t^i \quad (4)$$

2. Maximum keypoint position

Similar to the mean keypoint value, the maximum value of each keypoint can help distinguish between an overhead hand-raise movement for the "Stop" class and a near-head position for the "Thank & Greet" class. The value is calculated based on the following equation:

$$x_{max}^i = \max_t \hat{x}_t^i, \quad y_{max}^i = \max_t \hat{y}_t^i \quad (5)$$

3. Average distance between keypoints

Another static feature that can be extracted from the skeleton keypoints is the average distances between keypoints over the total frames, calculated using (6). For instance, the distance between the right and left hand may be longer for movements that include only one hand, and it may be shorter for simultaneous hand movements.

$$\bar{d}^{i,j} = \frac{1}{t}\sum_{t=0}^{t}\|\hat{\mathbf{p}}_t^i - \hat{\mathbf{p}}_t^j\| \tag{6}$$

Using the equations mentioned above, there are a total of 71 static features, comprising 17 $\bar{x}$ points, 17 $\bar{y}$ points, 17 $x_{max}$ points, 17 $y_{max}$ points, and 3 keypoint distances. We only calculated 3 keypoint distances, i.e., left hand (LW) to LS ($\bar{d}^{LW,LS}$), right hand (RH) to RS ($\bar{d}^{RW,RS}$), and LW to RW ($\bar{d}^{LW,RW}$). This selection is based on the dominant gestural movements found in the WIVW dataset, namely, by using the left and right hands.

### B. Dynamic Feature

While static features help to encode pose configuration, dynamic features are essential in differentiating movements with similar positions. For example, the hand position of "Stop" and "Thank & Greet" may have some similarities. While the "Stop" class may have less motion, the latter may include more changes in dynamic motions due to the presence of waving. Therefore, we extracted two dynamic feature descriptors: the velocity and acceleration of keypoints. Both features are calculated based on (7) and (8), consecutively.

$$\hat{\mathbf{v}}_t^i = \hat{\mathbf{p}}_t^i - \hat{\mathbf{p}}_{t-1}^i, \quad \bar{v}^i = \frac{1}{t}\sum_{t=0}^{t}\|\hat{\mathbf{v}}_t^i\| \tag{7}$$

$$\bar{a}^i = \frac{1}{t}\sum_{t=0}^{t}\|\hat{\mathbf{v}}_t^i - \hat{\mathbf{v}}_{t-1}^i\| \tag{8}$$

The normalised velocity of every $i$-th keypoint $\hat{\mathbf{v}}_t^i$ is calculated based on the difference between every normalised keypoint $\hat{\mathbf{p}}_t^i$ for every two consecutive frames ($t$ and $t-1$). These velocity values are then averaged to produce the velocity feature $\bar{v}^i$, whereas the acceleration feature $\bar{a}^i$ is the mean velocity value over the total sequence. We focused only on hands for both features because they are the dominant limbs in every gesture. There is an added velocity ratio for both hands to add as an extra feature, giving a total of 5 dynamic features, i.e., $\bar{v}^{LW}$, $\bar{v}^{RW}$, $\bar{v}^{LW}/\bar{v}^{RW}$, $\bar{a}^{LW}$, and $\bar{a}^{RW}$.

## IV. ANALYSIS AND RESULTS

This work aims to investigate the driving factors of pedestrian gestures and their influence on classifying the gestures. Therefore, we analysed the extracted features, obtained from the methods explained in Section III, using three feature subsets: static-only (71 features), dynamic-only (5 features), and their combined set (76 features).

In this work, we implemented the t-distributed Stochastic Neighbour Embedding (t-SNE) [27] method to visualise the clusters of each class. This method helps reduce the high number of dimensions of the static and dynamic features into two dimensions, which helps visualise the clusters of each gesture class and enhances the model explainablity. Accompanying each t-SNE cluster, a silhouette score is also calculated to help in comparing the quality of the clusters, as proposed by Rousseeuw [28].

We employed a classical machine learning method, the random forest classifier [29], to classify the videos using the three feature subsets as inputs. This method helps in providing more explainability of the model, instead of only focusing on the model's performance. Alzahrani et al. have proven that random forest is the best-performing algorithm, compared to three other supervised learning methods (C4.5, ANN, and SVM), in classifying skeleton features for activity recognition [25]. Additionally, we also provided a feature ranking to inspect the determining factor that distinguishes one gesture class from another. The rankings are provided based on the Gini importance score [30] that is widely used to rank input features from random forest classifications.

### A. Static Feature Analysis

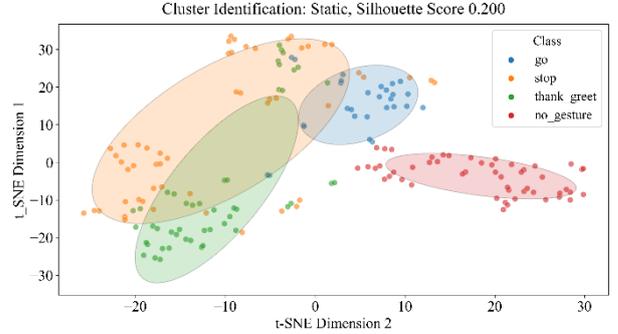

Fig. 5 t-SNE cluster identification of static features.

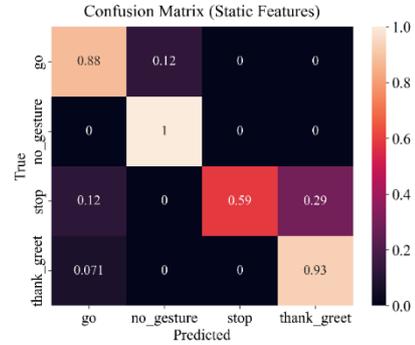

Fig. 6 Classification result of the static features.

Figure Fig. 5 shows the cluster identification based on the t-SNE evaluation of the static features. We utilised a Gaussian Mixture approach to localise the clusters, illustrated by the matching-colored circles in the figure. It is clearly shown that the "No Gesture" class is easily distinguishable from the other classes. This goes in line with the fact that there is minimal movement when the pedestrians are not gesturing. However, there is a similarity between the "Thank & Greet" class with the "Stop" class, as there are some overlapping areas between the two clusters. The similarity is likely caused due to an almost identical hand position between the two classes. In contrast, the "Go" class is almost separated from these two classes, which is very likely caused due to the lower hand and arm position during the arm-waving gesture.

The static features are then used as the input to classify the four gesture classes. The classification yielded an accuracy of 83.33%, with the corresponding confusion matrix shown in Fig. 6. It is clearly shown that the classifier has difficulty in detecting the "Stop" gesture. This is connected with the previous fact that some of the "Stop" class overlapped with the "Thank & Greet" class. The other classes have shown direct comparability with the t-SNE plot in Fig. 5, with perfect accuracy for the "No Gesture" class and 88% accuracy for the "Go" class.

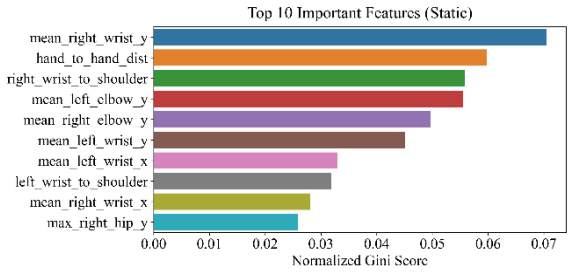

Fig. 7 Static feature ranking.

When the static features are compared, it is shown in Fig. 7 that the mean $y$ position of the right hand having the highest rank. This is likely caused by the pedestrians having their right hand stationary because it is most likely occluded from the vehicle's view. The second highest feature is the hand-to-hand distance, which is unsurprising because there is a very high chance of positional difference between the two hands when a person is gesturing.

*B. Dynamic Feature Analysis*

In comparison, the dynamic cluster in Fig. 8 shows that the "Thank & Greet" class, along with the "No Gesture" class, are almost entirely detached from the other clusters, except for a few outliers. The reason for this separation is likely caused by the quicker motions of the "Thank & Greet" class, as depicted also in Fig. 2. However, the t-SNE plot shows that the results of the dynamic gesture clustering are not as good as the static feature, with the latter having a higher silhouette score of 0.200 compared to 0.114 of the dynamic gesture.

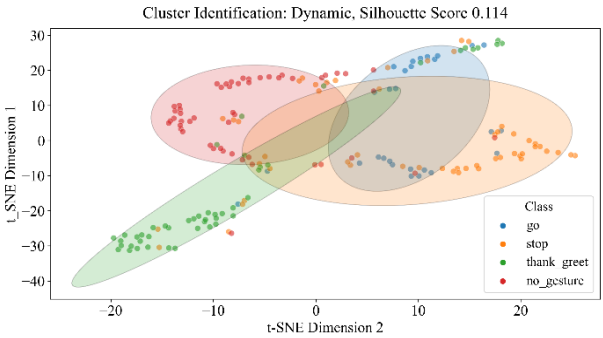

Fig. 8 t-SNE cluster identification of dynamic features.

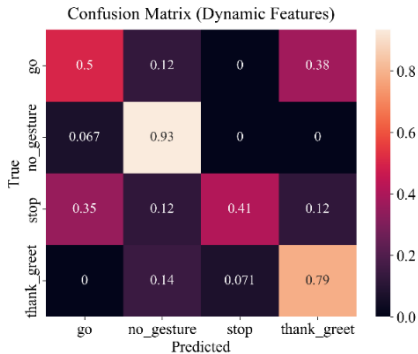

Fig. 9 Classification result of the dynamic features.

Due to the poor clustering shown in Fig. 8, the resulting classification using the dynamic gestures is not very satisfactory, only achieving an accuracy score of 66.67%. The detailed result of the classification is shown in Fig. 9.

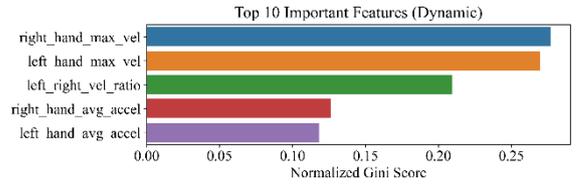

Fig. 10 Dynamic feature ranking.

The dynamic feature ranking in Fig. 10 shows that the maximum velocity of each hand plays a dominant role in determining the type of gesture. However, this could not be directly taken as the strongest contributing factor of the dynamic gestures until both features are combined and evaluated. The result of this process is detailed further in the next subsection.

*C. Combined Feature Analysis*

As shown in Fig. 11, the combination of both features improves the t-SNE clustering of the gesture classes, increasing the silhouette score from 0.200 to 0.239. Although the improvement seems insignificant, it can be seen that the introduction of dynamic gestures helps to moderately separate the "Stop" class and "Thank & Greet" class. However, it is evident that the "Go" class still has some similarity with the "Thank & Greet" class, which may be caused due to the similar gesture duration of both classes (see Fig. 2).

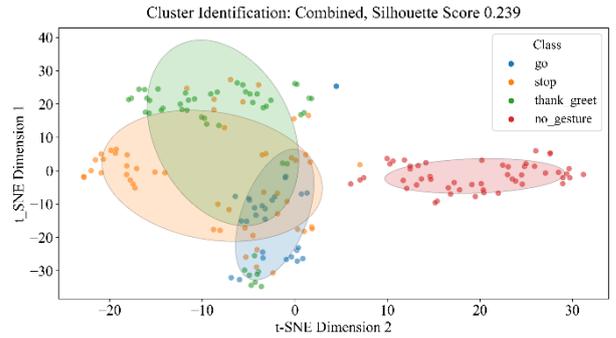

Fig. 11 t-SNE cluster identification of combined features.

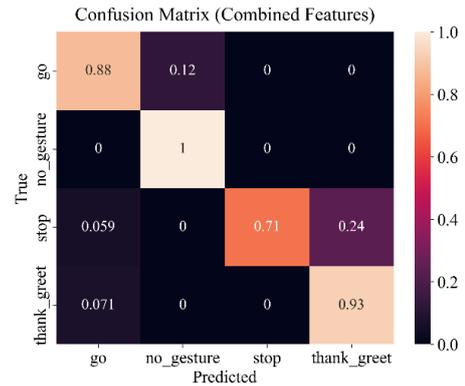

Fig. 12 Classification result of the combined features.

The gesture classification using the combined features improved the static-only classification, increasing the accuracy score from 83.33% to 87.04%. The resulting confusion matrix in Fig. 12 further emphasised the improvement found in the t-SNE evaluation, significantly increasing the "Stop" class accuracy from 59% to 71%. The errors found in the confusion matrix are likely caused due to some similar movements found in the "Stop" class that have periodic movements found in the "Go" and "Thank & Greet" classes.

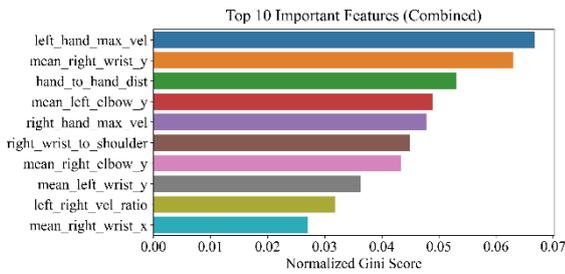

Fig. 13 Combined feature ranking.

Finally, the combined feature ranking in Fig. 13 shows that the velocity of the left hand has the highest contributing factor towards distinguishing the gestures. This is physically proven because when pedestrians are positioned on the right side of the road, they primarily use their left hand to gesture and signal toward vehicles. The second and third ranking feature shows that the right hand could serve as an anchor point. It could help in distinguishing gestures with longer hand-to-hand distances, as mentioned in the subsection IV.A.

## V. CONCLUSION AND FUTURE WORKS

This paper contributes a comprehensive framework for pedestrian gesture classification using real-world data, addressing one of the key challenges in pedestrian-AV interaction. Through careful selection of the WIVW dataset and defining a four-class taxonomy, we demonstrated that combining static and dynamic features achieves 87% accuracy. Our findings indicate that both static pose and motion dynamics are critical in differentiating gestures, with hand velocity and hand-to-hand distance serving as strong discriminators.

Future work can expand on these research findings by incorporating deep learning models or temporal sequence modelling (e.g., using LSTM). Other features that can be extracted from skeleton points, such as from the frequency domain could help in distinguishing the periodical movement difference. Moreover, adding a third feature, such as duration, can help separate the clusters that overlap in the t-SNE plots. These advances will be essential in developing an accurate and robust universal pedestrian gesture recognition.

ACKNOWLEDGMENT

This work is funded by the UK Research and Innovation (UKRI) Reference Numbers EP/W524372/1 and 2887445 through the EPSRC Doctoral Training Partnership Scheme. We would also like to thank Thomas Brand, Marcus Schmitz, and the WIVW for providing us access to the WIVW dataset.